\documentclass{article} 
\usepackage{arxiv,times}

\usepackage{amsmath,amsfonts,bm}

\def\eqref#1{equation~\ref{#1}}

\def\1{\bm{1}}

\DeclareMathAlphabet{\mathsfit}{\encodingdefault}{\sfdefault}{m}{sl}
\SetMathAlphabet{\mathsfit}{bold}{\encodingdefault}{\sfdefault}{bx}{n}

\usepackage{hyperref}
\usepackage{url}
\usepackage{graphicx}
\usepackage{wrapfig}
\usepackage{booktabs}
\usepackage{amssymb}
\usepackage{algorithm}
\usepackage{algpseudocode}

\title{DroneWAM: Efficient World Action Model for Drone Visual Navigation}

\author{
Liang Yao$^{1}$ \quad
Fan Liu$^{1,2,}$ \thanks{Corresponding authors.} \quad
Hongbo Lu$^{3,7}$ \quad
Wei Xu$^{4}$ \quad
Jianyu Jiang$^{5}$ \quad
Yijun Shen$^{6}$ \quad \\
\textbf{Chuanyi Zhang}$^{1}$ \quad
\textbf{Pai Peng}$^{7,}$ \footnotemark[1]
\\
{\small
$^{1}$Hohai University \quad 
$^{2}$Key Laboratory of Water Big Data Technology of Ministry of Water Resources \quad
}
\\
{\small
$^{3}$Shanghai Jiao Tong University \quad
$^{4}$Nanjing University
$^{5}$Southeast University \quad
}
\\
{\small
$^{6}$East China Normal University \quad
$^{7}$CowaRobot Co. Ltd.
}
}

\iclrfinalcopy 
\begin{document}

\maketitle

\begin{abstract}
World-action models give visual navigation agents a way to anticipate how candidate actions will change future observations and to act from the predicted consequences. 
For drones, this capability must operate under tight accuracy and efficiency constraints. We present DroneWAM, an efficient world-action model for drone visual navigation. DroneWAM adopts a JEPA-based architecture to model future states directly in representation space, avoiding the cost of explicit future image generation. A pretrained Resampler further compresses dense encoder features into fewer latent tokens, reducing the computation repeated at each imagined step. We also introduce adaptive rollout, where a preference-trained Gate adaptively allocates prediction depth according to the current scene. To support learning under richer aerial motion, we construct DroneNav-6D, a simulated visual navigation dataset with synchronized RGB observations, 6-DoF flight trajectories, control commands, and randomized wind disturbances. On DroneNav-6D, DroneWAM achieves the best trajectory accuracy among the compared methods.
Adaptive rollout further reduces the average prediction depth from 8 to 4.58 while improving trajectory accuracy, demonstrating that predictive computation can be allocated more effectively across scenes. \href{https://github.com/1e12Leon/DroneWAM}{Codes and data} will be released.
\end{abstract}

\section{Introduction}
Visual navigation~\citep{zhang2022survey,nahavandi2025comprehensive,jiang2026airnavigation} is a central capability for autonomous drones. A drone navigates from partial egocentric observations, while each control command changes both its physical state and future visual input. Therefore, reliable navigation requires anticipating how candidate actions may affect future observations and progress toward the goal. World Models~\citep{matsuo2022deep,chen2025planning} provide such predictive capability by modeling state transitions, while World-Action Models (WAMs)~\citep{shen2026world,lu2026dawn,lu2026rise} further connect future prediction with action generation. Since this prediction is repeatedly invoked throughout closed-loop flight under limited onboard computation, an aerial WAM should provide accurate foresight with high inference efficiency.

\begin{wrapfigure}{r}{0.60\textwidth}
    \centering
    \vspace{-0.5cm}
    \includegraphics[width=0.99\linewidth]{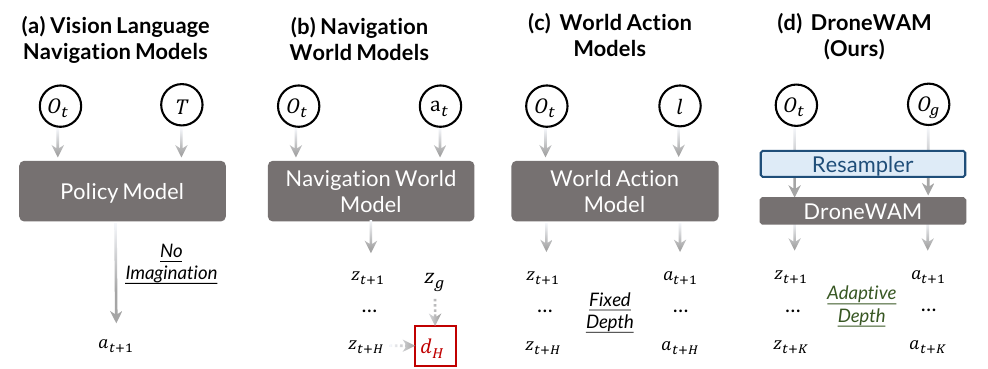}
    \vspace{-0.3cm}
    \caption{Compared to prior approaches.}
    \vspace{-0.3cm}
    \label{teaser}
\end{wrapfigure}

\looseness=-1
Existing aerial navigation methods mainly approach this problem from two directions. As shown in Fig.~\ref{teaser}, vision-language navigation methods~\citep{liu2023aerialvln} condition action prediction on visual observations and semantic instructions, providing an effective interface for goal-directed navigation when language supervision is available. However, their future visual consequences are usually modeled only implicitly. Aerial world models and WAMs~\citep{zhang2025aerial,zhao2026worldvln,zheng2026worldfly} instead explicitly predict how the visual world evolves with motion and use the predicted future to support action generation. Such predictive reasoning introduces additional computation during online control, since visual representations must be repeatedly propagated over multiple imagined steps. Moreover, existing models commonly use a predefined rollout horizon, assigning similar predictive computation to different scenes.

These limitations suggest that efficiency should be considered directly in the predictive process of an aerial WAM. We identify three aspects that are particularly important. Firstly, the prediction space should retain the scene structure and motion information required for navigation without incurring the cost of unnecessary visual details. Secondly, the representation propagated through the world model should be compact, since its cost is repeatedly accumulated across rollout steps. Thirdly, the amount of prediction should depend on the current scene: straightforward situations may require only limited foresight, while ambiguous observations can benefit from deeper rollout. Together, these considerations motivate a world-action model that is both predictive and economical in how it represents and imagines the future.

Furthermore, while recent efforts have expanded the scale of video-action datasets~\citep{chen2026action100m}, training data for aerial world-action models should also capture the diverse translational and rotational motion of drones. Most established aerial visual-navigation benchmarks~\citep{liu2023aerialvln,zhu2026uncertainty,wang2025towards,yao2025uemm} represent flight with 4-DoF motion over translation and yaw. This abstraction provides limited coverage of roll and pitch, whose changes directly affect camera viewpoint and visual dynamics during flight. Higher-DoF visual-action data are therefore desirable for learning a more complete aerial world-action model.

Guided by these observations, we present DroneWAM, an efficient world-action model for drone visual navigation. DroneWAM adopts JEPA-based predictive learning~\citep{assran2025v,assran2023self,VL-JEPA} to model future states directly in representation space, focusing computation on predictive scene structure and motion cues that are most relevant to navigation. A pretrained Resampler further compresses dense visual features into a compact token set, reducing the computation required at each imagined step. DroneWAM also adapts its rollout depth to the current scene through a preference-trained~\citep{rafailov2023direct} Gate, allowing straightforward situations to use fewer prediction steps while allocating additional foresight to more ambiguous ones. These designs reduce both the spatial and temporal cost of world-action prediction while preserving the predictive information needed for navigation.

To support learning under richer aerial motion, we further construct DroneNav-6D, a simulated dataset for 6-DoF visual navigation. 
Since collecting large-scale real 6-DoF flight data is costly, we develop an automatic simulation pipeline that records synchronized RGB observations, flight commands, and 6-DoF trajectories across diverse environments, with randomized wind disturbances to broaden the covered flight conditions. DroneNav-6D provides more complete motion supervision for training world-action models and enables controlled evaluation under full 6-DoF aerial motion.

We evaluate DroneWAM on DroneNav-6D under both open-loop prediction and closed-loop navigation. In open-loop evaluation, DroneWAM achieves the lowest trajectory errors among all compared methods, reducing ATE by 39.8\% and inference latency by 32.2\% compared with FastWAM. In closed-loop evaluation, it improves navigation progress from 0.510 to 0.774 while maintaining lower trajectory error and inference cost. Ablation studies further validate the efficiency-oriented design: compressing the visual representation to 128 tokens reduces inference time by 18.3\% compared with a 512-token representation with nearly unchanged trajectory accuracy, while adaptive rollout reduces the average prediction depth from 8 to 4.58 and lowers latency by 22.4\% over fixed-depth rollout while achieving lower trajectory errors. We further observe similar benefits on LIBERO~\citep{liu2023libero}, where adaptive rollout reduces inference latency by 53.8\% relative to fixed-depth prediction with slightly improved action prediction accuracy. Our contributions are as follows:
\begin{itemize}
    \item We introduce DroneWAM, an efficient world-action model for drone visual navigation. It provides a practical modeling framework for predictive navigation under constrained onboard computation.

    \item We construct DroneNav-6D, a simulated 6-DoF aerial navigation dataset built with an automatic data generation pipeline. It offers a scalable resource for training and evaluating aerial world-action models beyond conventional 4-DoF motion.

    \item Extensive experiments demonstrate that DroneWAM achieves both higher navigation accuracy and lower inference cost than existing world-model and world-action baselines. 
\end{itemize}

\section{DroneNav-6D}
\label{sec:dronenav6d}

To support predictive learning with translational and attitude changes, we develop an automatic data collection pipeline using Unreal Engine 4.27 and AirSim~\citep{shah2017airsim}. AirSim provides multirotor simulation, RGB rendering, and synchronized state recording. We collect data in five simulation worlds: traffic roads in SnappyRoads, snowy mountains, a coastal city in CITYBIM, rural and agricultural areas in RuralAustralia, and dense urban streets in NYC. These worlds provide varied scene layouts, terrain, and visual appearances.

\begin{wrapfigure}{R}{0.56\textwidth}
  \centering
  \vspace{-0.4cm}
  \includegraphics[width=\linewidth]{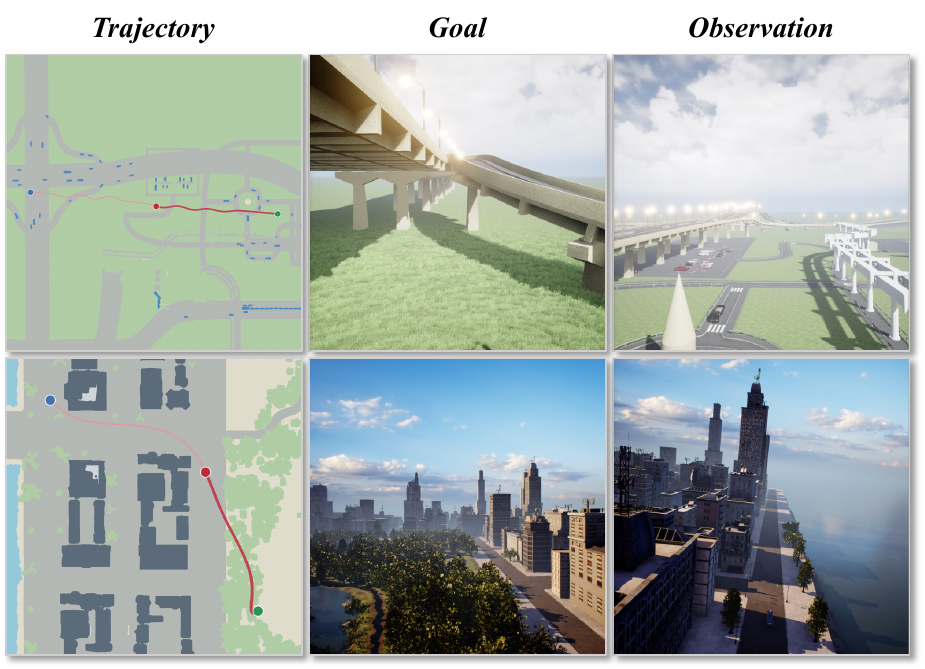}
  \caption{Examples of goal-conditioned navigation samples in DroneNav-6D. Each row shows the flight trajectory together with the sampled goal image and current observation.}
  \label{fig:dronenav_goal}
\end{wrapfigure}

For each flight, a multirotor equipped with a forward-facing RGB camera ascends to a predefined cruising altitude. We sample waypoints within scene-specific flight boundaries by varying the travel direction, displacement, and altitude offset. The waypoints are connected into a smooth three-dimensional trajectory and executed through velocity and heading commands. During execution, we record RGB observations together with the vehicle position, orientation, and issued commands. Consecutive records associate visual transitions with commanded inputs and realized pose changes, providing supervision for future prediction.

We inject random wind disturbances during flight and record the resulting motion, including lateral drift, attitude changes, and controller corrections. The paired command and pose records capture the commanded inputs and realized motion under these disturbances. Episodes are filtered using checks on collisions, prolonged immobility, trajectory length, frame intervals, and action validity. This procedure yields temporally ordered visual and motion sequences for predictive learning.

The resulting DroneNav-6D corpus contains 3,837 trajectory clips with 1,592,170 RGB observations, totaling 63.03 hours of recorded flight. The median recording rate is approximately 7.00\,Hz. Training and validation contain 3,453 and 384 clips, respectively, with distinct clip identifiers and all five worlds represented in both splits.

To support image-goal navigation, we further construct goal-conditioned samples from the recorded flight trajectories. For each sample, the current observation and a future goal image are selected from the same trajectory, while the synchronized intermediate states and actions provide the corresponding motion supervision. As illustrated in Fig.~\ref{fig:dronenav_goal}, we show two representative navigation samples from DroneNav-6D. The first corresponds to a relatively open environment with an approximately straight navigation trajectory, while the second is collected in a dense urban area where the drone needs to turn between buildings, representing a more complex navigation case. Further dataset statistics and analyses are provided in Appendix~\ref{app:dataset_analysis}.


\section{DroneWAM}

\subsection{Problem Setup}

Following prior work on image-goal navigation~\citep{shah2023vint,sridhar2024nomad}, we specify the navigation target with a goal image $g$. At control step $t$, the drone receives an RGB observation $o_t$ and a motion state $s_t$, and has access to its executed flight commands $a_{\leq t}$. The observation and goal image are processed by the same visual encoder $E$ and Resampler $R_\rho$, yielding $z_t = R_\rho(E(o_t))$ and $z_g = R_\rho(E(g))$, respectively. Given the observed context and the goal latent $z_g$, a world-action model predicts future latent states and an $H$-step action plan for navigation. We use $k$ to denote the number of imagined world transitions and $h$ to denote the temporal index within an action plan. Thus, the rollout depth and the action horizon describe two different temporal dimensions.

Conventional WAMs perform a predefined number of world transitions $K$ for each control update. We instead formulate world-action prediction as a variable-length rollout:
\begin{equation}
    \tau_t
    =
    \min\left(
        \left\{k \mid d_{t,k} = \mathrm{stop}\right\}
        \cup
        \left\{K_{\max}\right\}
    \right),
\end{equation}
where $d_{t,k}$ is the stopping decision after the $k$-th imagined transition. At each depth $k$, the model produces a predicted latent state $\hat{z}_{t+k}$ and an $H$-step action plan $\hat{\mathbf{A}}_t^{(k)}$. The resulting depth $\tau_t$ therefore controls the amount of world prediction performed for the current scene, while $H$ remains the planning horizon of each candidate plan.



\begin{figure}[t]
    \centering
    \includegraphics[width=0.99\linewidth]{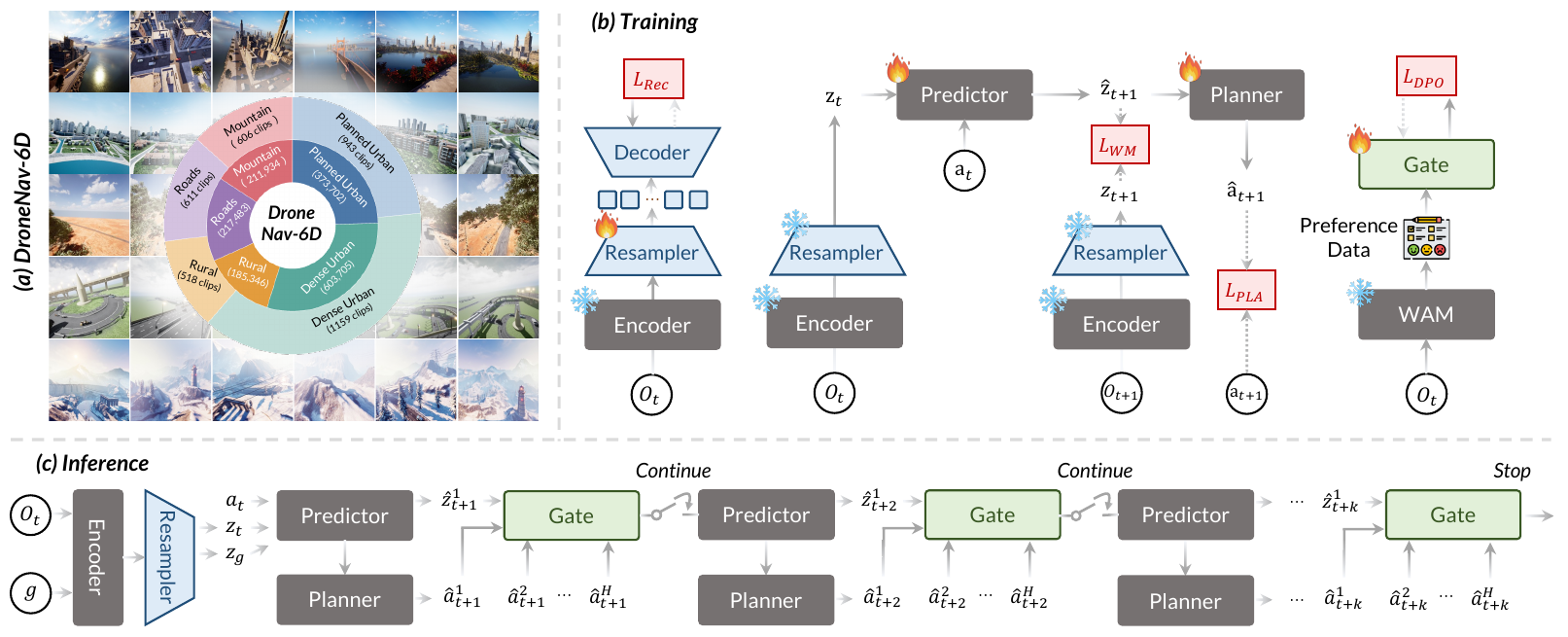}
    \caption{
        Overview of DroneWAM and DroneNav-6D.
        (a) DroneNav-6D provides diverse simulated environments with synchronized visual observations and 6-DoF flight trajectories.
        (b) DroneWAM is trained in three stages: Resampler pretraining, world-action model training with latent prediction and planning objectives, and preference learning for the Gate.
        (c) During inference, the Predictor and Planner iteratively imagine future latent states and action plans, while the Gate determines when to stop and returns the highest-scoring plan for efficiency.
        }
    \label{overview}
\end{figure}



\subsection{Training}

\paragraph{Resampler pretraining.}
Given an RGB observation $o_t$, a frozen visual encoder $E$ produces dense visual tokens $x_t = E(o_t) \in \mathbb{R}^{M \times d}$. We pretrain a lightweight Resampler $R_\rho$ to compress them into a fixed set of $N = 128$ latent tokens, $z_t = R_\rho(x_t)$, while a lightweight Decoder reconstructs the original encoder features from the compressed representation. The Resampler and Decoder are trained with feature reconstruction and temporal consistency objectives. This produces a compact visual representation that preserves the encoder information while reducing the computation of subsequent multi-step prediction. After pretraining, the encoder and Resampler are frozen, and the Decoder is discarded. The goal image is encoded using the same frozen encoder and Resampler to obtain the goal latent $z_g = R_\rho(E(g))$.

\paragraph{Predictor and Planner training.}
Let $C_t = \{z_{\leq t}, s_{\leq t}, a_{\leq t}\}$ denote the observed visual, motion-state, and executed-action context. Conditioned on $C_t$ and the goal latent $z_g$, the Predictor $F_\theta$ autoregressively predicts up to $K_{\max}$ future latent states:
\begin{equation}
    \hat{z}_{t+k}
    =
    F_\theta\left(
        C_t,
        z_g,
        \hat{z}_{t+1:t+k-1}
    \right),
    \qquad
    k = 1,\ldots,K_{\max}.
\end{equation}

Future ground-truth states and actions are masked from the Predictor input during training. The Predictor therefore uses only the observed context, the goal latent, and previously predicted latent states. 
We denote the target representation at depth $k$ by $z^\star_{t+k}$ and supervise world prediction with a temporally discounted latent loss:
\begin{equation}
    \mathcal{L}_{\mathrm{WM}}
    =
    \frac{
        \sum_{k=1}^{K_{\max}}
        \gamma^{k-1}
        \left\lVert
            \hat{z}_{t+k} - z^\star_{t+k}
        \right\rVert_1
    }{
        \sum_{k=1}^{K_{\max}} \gamma^{k-1}
    }.
\end{equation}

At each imagination depth $k$, the Planner $P_\psi$ takes the predicted latent prefix and produces a complete $H$-step 6-DoF action plan~\citep{zhao2023learning} :
\begin{equation}
    \hat{\mathbf{A}}_t^{(k)}
    =
    P_\psi\left(\hat{z}_{t+1:t+k}, z_g\right)
    =
    \left(
        \hat{a}_{t+1\mid t}^{(k)},
        \ldots,
        \hat{a}_{t+H\mid t}^{(k)}
    \right).
\end{equation}

Here, $k$ denotes the world imagination depth, while $h \in \{1,\ldots,H\}$ indexes the temporal position within the action plan. Thus, the $H$ outputs form one future action sequence rather than $H$ candidate actions at the same time step.

Each action plan is integrated from the current 6-DoF pose to obtain a predicted trajectory. We supervise the Planner at every imagination depth using relative absolute trajectory error (rATE) and relative pose error (rRPE):
\begin{equation}
    \begin{aligned}
        \mathcal{L}_{\mathrm{PLA}}
        &=
        \frac{1}{K_{\max}}
        \sum_{k=1}^{K_{\max}}
        \left(
            \frac{\mathrm{rATE}^{(k)}}{\epsilon_A}
            +
            \frac{\mathrm{rRPE}^{(k)}}{\epsilon_R}
        \right), \\
        \mathcal{L}_{\mathrm{WAM}}
        &=
        \lambda_{\mathrm{WM}} \mathcal{L}_{\mathrm{WM}}
        +
        \lambda_{\mathrm{PLA}} \mathcal{L}_{\mathrm{PLA}}.
    \end{aligned}
\end{equation}

Only the Predictor and Planner are optimized in this stage, while the representation modules and target branch remain frozen.

\paragraph{Gate preference training.}
After WAM training, we freeze the complete WAM and evaluate every imagination depth $k \in \{1,\ldots,K_{\max}\}$ for each sample $i$. Each depth produces an $H$-step action plan and its corresponding trajectory errors. We define a cost that jointly considers planning quality and rollout computation:
\begin{equation}
    C_{i,k}
    =
    \frac{\mathrm{rATE}_{i,k}}{\epsilon_A}
    +
    \frac{\mathrm{rRPE}_{i,k}}{\epsilon_R}
    +
    \lambda_c \frac{k}{K_{\max}}.
\end{equation}
The depth with the lowest cost is treated as the preferred sample, while the remaining depths form rejected samples.

The Gate input $u_{i,k}$ contains only information available at inference time, including the observed latent state, predicted latent history, action-plan history, changes between adjacent predictions, recent 6-DoF motion, current velocity, and the imagination depth. The Gate predicts low-reward and high-reward logits. Given a fixed reference Gate, we define a DPO-style reference-adjusted reward and optimize the Gate:
\begin{equation}
    \begin{aligned}
        r_\omega(u)
        &=
        \log
        \frac{
            p_\omega(\mathrm{high}\mid u)
        }{
            p_\omega(\mathrm{low}\mid u)
        }
        -
        \log
        \frac{
            p_{\mathrm{ref}}(\mathrm{high}\mid u)
        }{
            p_{\mathrm{ref}}(\mathrm{low}\mid u)
        }, \\
        \mathcal{L}_{\mathrm{DPO}}
        &=
        -\mathbb{E}
        \left[
            \log \sigma
            \left(
                \beta
                \left[
                    r_\omega(u^+) - r_\omega(u^-)
                \right]
            \right)
        \right].
    \end{aligned}
\end{equation}
Only the Gate is updated during this stage. We then use a disjoint calibration subset to determine a stopping threshold $\eta_k$ for each imagination depth.

\subsection{Adaptive Inference}

At control step $t$, the shared frozen Encoder and Resampler encode the observed images and the goal image into latent representations. Conditioned on the observed context $C_t$ and the goal latent $z_g$, the model then performs world prediction, planning, and Gate evaluation iteratively. At depth $k$, the Predictor generates $\hat{z}_{t+k}$, the Planner produces $\hat{\mathbf{A}}_t^{(k)}$, and the Gate assigns the current plan a high-reward score $q_{t,k}$. The rollout stops once the score exceeds the calibrated threshold:
\begin{equation}
    \begin{aligned}
        \tau_t
        &=
        \min\left(
            \left\{
                k \mid
                K_{\min} \leq k \leq K_{\max},
                \ q_{t,k} \geq \eta_k
            \right\}
            \cup
            \left\{K_{\max}\right\}
        \right), \\
        \kappa_t
        &=
        \operatorname*{arg\,max}_{1 \leq k \leq \tau_t}
        q_{t,k}.
    \end{aligned}
\end{equation}

The stopping depth $\tau_t$ determines the number of world-prediction steps actually performed, while $\kappa_t$ determines which action plan is finally selected. The two may differ because the model retains the highest-scoring plan encountered before termination. The controller executes only the first action of $\hat{\mathbf{A}}_t^{(\kappa_t)}$. After receiving the next observation, the entire procedure is repeated in a receding-horizon manner. The goal image is provided as task input.

\section{Experiments}
\label{sec:experiments}

\subsection{Experimental Setup}

\paragraph{Implementation Details.}
DroneWAM predicts an eight-step ego-frame 6-DoF action plan from four RGB observations, associated 7-D motion states, executed-action history, and a goal image. Observations and goals share a frozen ViT-L encoder and pretrained TokenAE resampler ($512 \rightarrow 128$ tokens), with goal latents supplied at every imagination depth. The Predictor has 24 Transformer layers of width 1024, and the Planner maintains a 256-D recurrent state. Base WAM training uses 30 epochs of AdamW with bfloat16, a peak learning rate of $7.5 \times 10^{-5}$, and an effective batch size of 128, followed by five epochs of variable-depth Predictor--Planner adaptation. The five-member marginal-value Gate is trained for 30 epochs. Model training uses 8 $\times$ A100-80GB GPUs.

\paragraph{Metrics.}
We report Absolute Trajectory Error (ATE), Relative Pose Error (RPE), and their trajectory-length-normalized variants, rel.~ATE and rel.~RPE. We additionally report synchronized inference latency and average autoregressive rollout depth. Closed-loop evaluation further includes task progress, and MSE.

\subsection{Main Results}
\label{sec:main_results}

\textbf{Open-Loop Results.}
We first evaluate prediction accuracy and inference efficiency under open-loop rollout, where all methods are given the same observation context and predict future motion without intermediate feedback. As shown in Tab.~\ref{tab:open_loop}, DroneWAM achieves the best performance across all trajectory metrics while also requiring the lowest inference latency. Compared with FastWAM, DroneWAM reduces ATE from 1.8366 to 1.1062 and Rel.ATE from 0.1919 to 0.0578, while reducing inference time from 712.33\,ms to 483.29\,ms. 
These results show that DroneWAM achieves a better accuracy--efficiency trade-off than the compared baselines, with lower trajectory error and lower inference latency.

\begin{table}[t]
\centering
\setlength{\tabcolsep}{4.0pt}
\caption{Open-loop evaluation on DroneNav-6D.}
\label{tab:open_loop}
\begin{tabular}{l|cccc|r}
\toprule
Method & ATE$\downarrow$ & Rel.ATE$\downarrow$ & RPE$\downarrow$ &
Rel.RPE$\downarrow$ & Infer. (ms)$\downarrow$ \\
\midrule
DINO-WM~\citep{zhou2024dino}     & 9.0923  & 0.4459 & 3.0681 & 0.1502 & 29,622.24 \\
V-JEPA 2-AC~\citep{assran2025v} & 27.4317 & 0.6621 & 9.0439 & 0.2190 & 64,617.72 \\
LeWM~\citep{maes2026lewm}        & 4.2259  & 0.4164 & 1.0924 & 0.1080 & 654.37 \\
FastWAM~\citep{yuan2026fast}     & 1.8366  & 0.1919 & 0.5086 & 0.0427 & 712.33 \\
NWM~\citep{bar2025navigation}         & 3.7580  & 0.3740 & 0.8051 & 0.0802 & 403,827.78 \\
\midrule
Ours        & \textbf{1.1062} & \textbf{0.0578} &
\textbf{0.3470} & \textbf{0.0180} & \textbf{483.29} \\
\bottomrule
\end{tabular}
\end{table}

\textbf{Closed-Loop Results.}
We further evaluate closed-loop navigation in Tab.~\ref{tab:closed_loop}, where the model replans from new observations. DroneWAM achieves the highest progress of 0.774, the lowest trajectory errors and MSE, and lower inference latency than the compared methods. However, none of the evaluated methods completes the task, so the current system is not yet ready for real-world deployment. These gains nevertheless suggest that efficient world-action modeling is a promising direction for practical drone navigation. Additional closed-loop evaluations are provided in Appendix~\ref{app:additional-experiments}.


\begin{table}[t]
\centering
\caption{Closed-loop navigation results on DroneNav-6D.}
\label{tab:closed_loop}
\resizebox{\linewidth}{!}{%
\begin{tabular}{l|cccc|cc|r}
\toprule
Method & ATE$\downarrow$ & Rel.ATE$\downarrow$ & RPE$\downarrow$ &
Rel.RPE$\downarrow$ & Prog.$\uparrow$ & MSE$\downarrow$ &
Infer. (ms)$\downarrow$ \\
\midrule
FastWAM~\citep{yuan2026fast} & 119.8063 & 0.7410 & 2.7309 & 0.0176 & 0.510 & 124.30 & 712.33 \\
NWM~\citep{bar2025navigation}    & 88.6305  & 0.5980 & 2.7245 & 0.0171 & 0.500 & 178.55 & 403,827.78 \\
\midrule
Ours    & \textbf{83.4461} & \textbf{0.5184} &
\textbf{2.7194} & \textbf{0.0168} & \textbf{0.774} &
\textbf{91.96} & \textbf{483.29} \\
\bottomrule
\end{tabular}
}
\end{table}

\subsection{Ablation Studies}
\label{sec:ablation}

\subsubsection{Component Ablation}

\begin{wraptable}{r}{0.60\textwidth}
\centering
\vspace{-0.3cm}
\caption{Component ablation. PLA, RS, and AR denote Planner, Resampler, and Adaptive Rollout, respectively.}
\label{tab:component_ablation}
\begin{tabular}{ccc|cc|r}
\toprule
PLA & RS & AR & Rel.ATE$\downarrow$ & Rel.RPE$\downarrow$ &  Infer. (ms)$\downarrow$ \\
\midrule
           &            &            & 0.6621 & 0.2190 & 64,617.72 \\
\checkmark &            &            & 0.0584 & 0.0189 & 656.12  \\
\checkmark & \checkmark &            & 0.0603 & 0.0192 & 582.65  \\
\checkmark & \checkmark & \checkmark & \textbf{0.0578} &
\textbf{0.0180} & \textbf{483.29} \\
\bottomrule
\end{tabular}
\vspace{-0.2cm}
\end{wraptable}

We progressively introduce the major components to examine their contributions to prediction accuracy and efficiency. The planner provides the main improvement in trajectory accuracy. Adding the Resampler reduces inference time from 656.12\,ms to 582.65\,ms with only marginal changes in trajectory error, showing that dense encoder features contain substantial redundancy for world prediction. Adaptive Rollout further reduces latency to 483.29\,ms while improving both Rel.ATE and Rel.RPE. The Resampler and Adaptive Rollout reduce computation from complementary spatial and temporal perspectives.

\subsubsection{Adaptive Rollout Strategy}

\begin{wraptable}{r}{0.53\textwidth}
\centering
\vspace{-0.65cm}
\caption{Comparison of rollout strategies. Avg.~$K$ denotes the average rollout depth.}
\label{tab:rollout_strategy}
\begin{tabular}{l|ccc}
\toprule
Method & Rel.ATE$\downarrow$ & Rel.RPE$\downarrow$ & Avg.~$K$$\downarrow$ \\
\midrule
Fixed         & 0.0688 & 0.0221 & 8.000 \\
Random        & 0.0722 & 0.0224 & 4.581 \\
Latent Margin & 0.0887 & 0.0309 & 5.625 \\
\midrule
Ours          & \textbf{0.0578} & \textbf{0.0180} & \textbf{4.578} \\
\bottomrule
\end{tabular}
\end{wraptable}

We compare different stopping strategies to determine whether the gain arises from early termination itself or from selecting an appropriate rollout depth. Our method reduces the average rollout from 8.0 to 4.578 while achieving lower trajectory errors than fixed rollout. More importantly, random stopping uses nearly identical computation with an average depth of 4.581, yet performs considerably worse. This shows that the learned Gate improves the allocation of predictive computation rather than simply shortening the rollout.

\subsubsection{Number of Resampler Tokens}

\begin{wraptable}{r}{0.48\textwidth}
\centering
\vspace{-0.6cm}
\caption{Effect of Resampler size. \#Tok. denotes the number of output latent tokens.}
\label{tab:resampler_tokens}
\begin{tabular}{c|ccc}
\toprule
\#Tok. & Rel.ATE$\downarrow$ & Rel.RPE$\downarrow$ &  Infer. (ms)$\downarrow$ \\
\midrule
512 & \textbf{0.0574} & \textbf{0.0179} & 591.28 \\
128 & 0.0578 & 0.0180 & 483.29 \\
16  & 0.0659 & 0.0198 & \textbf{405.16} \\
\bottomrule
\end{tabular}
\end{wraptable}

We vary the number of Resampler tokens to study the trade-off between representation capacity and inference efficiency. Compressing 512 tokens to 128 reduces latency from 591.28\,ms to 483.29\,ms with almost unchanged Rel.ATE and Rel.RPE. This yields an 18.3\% latency reduction with absolute error increases of only 0.0004 and 0.0001, respectively. Further compression to 16 tokens reduces latency again but causes a clear degradation in trajectory accuracy. We therefore use 128 tokens as the default configuration, providing a favorable balance between compactness and predictive fidelity.

\subsection{Further Analysis}
\label{sec:further_analysis}

\paragraph{Why Does Adaptive Rollout Improve Prediction Accuracy?}

\begin{wrapfigure}{r}{0.6\textwidth}
    \centering
    \vspace{-0.6cm}
    \includegraphics[width=\linewidth]{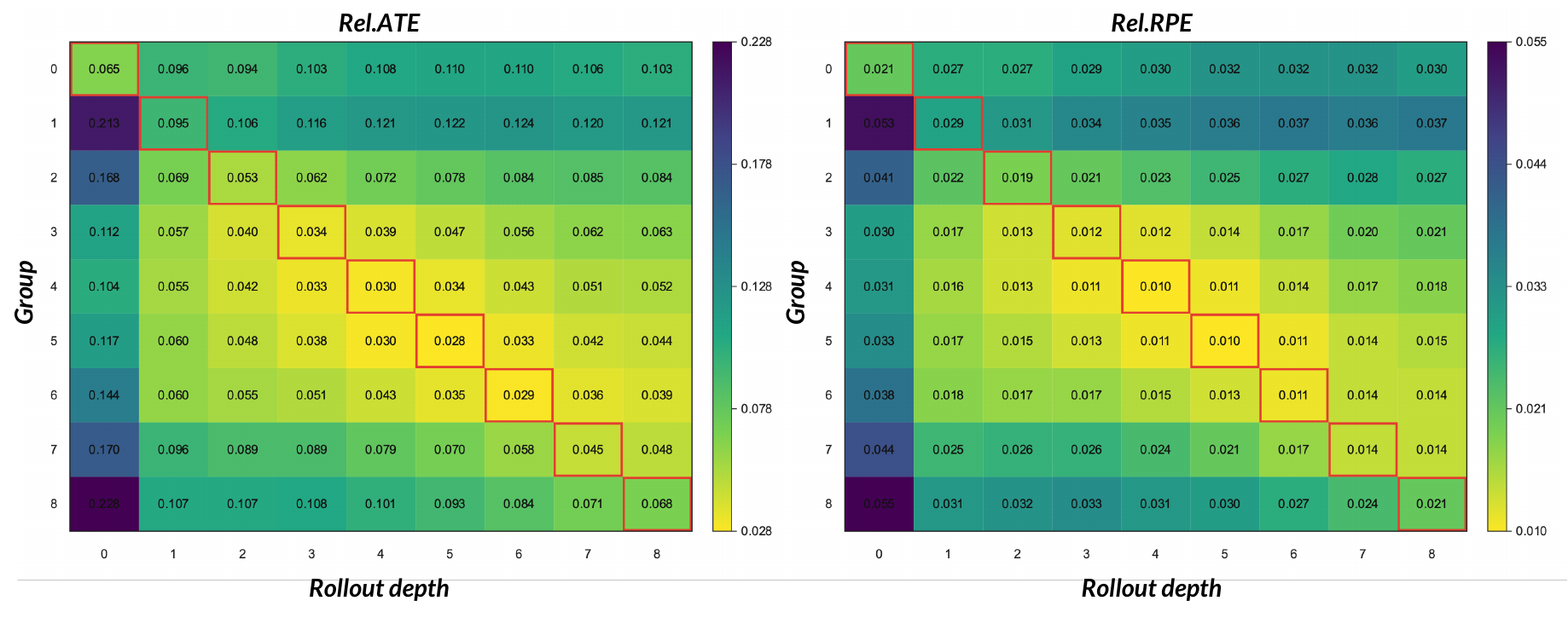}
    \vspace{-0.6cm}
    \caption{Accuracy at different rollout depths. Samples are grouped by the rollout depth, and red boxes indicate the corresponding selected depth.}
    \label{fig:rollout_heatmap}
\end{wrapfigure}

We evaluate every sample at all rollout depths from $K=0$ to $K=8$ and group it by the depth that achieves the lowest trajectory error. As shown in Fig.~\ref{fig:rollout_heatmap}, the optimal depths vary substantially across samples rather than concentrating at the maximum rollout depth. Some samples are best predicted with only shallow imagination, while others continue to benefit from deeper rollout, indicating that more prediction is not always better. It suggests that the useful amount of imagination depends on the current scene. Adaptive rollout can therefore improve both accuracy and efficiency by learning to allocate an appropriate prediction depth to each input.

\paragraph{Does Adaptive Rollout Generalize Beyond Aerial Navigation?}

\begin{wraptable}{r}{0.55\textwidth}
\centering
\vspace{-0.4cm}
\caption{Open-loop results on LIBERO. Full action metrics are reported in the Appendix.}
\label{tab:libero}
\begin{tabular}{lccc}
\toprule
Method & N.RMSE$\downarrow$ & Grip.(\%)$\uparrow$ & Latency$\downarrow$ \\
\midrule
$K=8$ & 0.2685 & 89.83 & 890.8 \\
Adaptive &
\textbf{0.2652} &
\textbf{90.35} &
\textbf{411.9} \\
\bottomrule
\end{tabular}
\end{wraptable}

To examine whether adaptive computation is specific to aerial navigation, we apply the same rollout strategy to LIBERO robotic manipulation. Adaptive rollout reduces inference latency from 890.8\,ms to 411.9\,ms, a 53.8\% reduction, while slightly improving normalized action error and gripper accuracy. The remaining action-error metrics show the same trend and are reported in the Appendix. Moreover, the improvement over fixed $K=8$ holds across all four LIBERO suites, while the average selected depth varies across task families. These results suggest that adaptive rollout transfers beyond aerial navigation and adjusts its computation to different task distributions.

\paragraph{Can the Preference-Trained Gate Generalize to Unseen Environments?}

\begin{wrapfigure}{r}{0.50\textwidth}
    \centering
    \vspace{-0.5cm}
    \includegraphics[width=\linewidth]{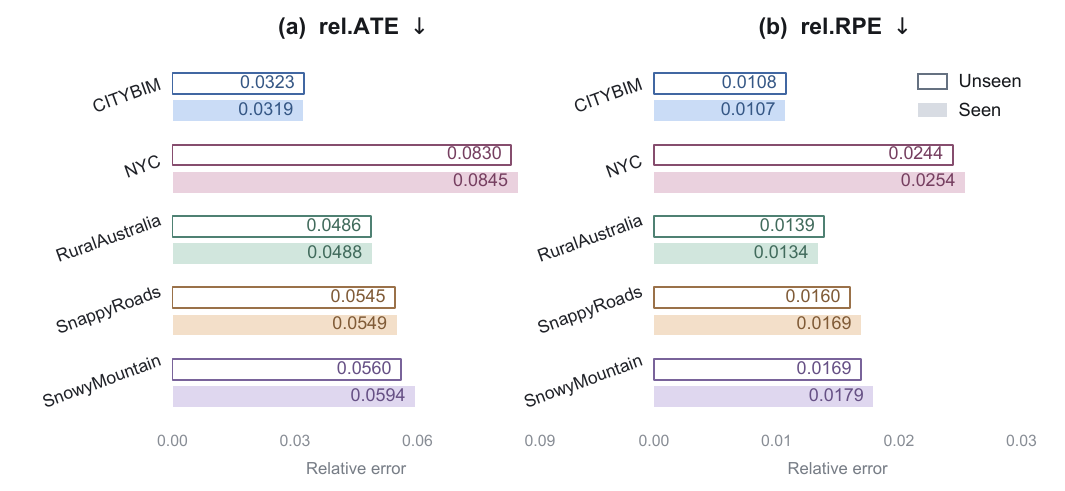}
    \vspace{-0.5cm}
    \caption{Generalization of the preference-trained Gate across unseen environments. Seen and unseen results are compared under environment-level cross-validation.}
    \label{fig:gate_generalization}
\end{wrapfigure}


We further examine whether preference learning causes the Gate to overfit to environments observed during training. We use environment-level cross-validation, defining environments included in Gate preference training as seen and those held out from this training as unseen. As shown in Fig.~\ref{fig:gate_generalization}, the seen and unseen results remain closely aligned for both Rel.ATE and Rel.RPE across all five environments. The maximum absolute gap is only 0.0034 for Rel.ATE and 0.0010 for Rel.RPE, and unseen errors are lower in seven of the ten environment--metric comparisons. Importantly, each unseen result is obtained without preference pairs from the corresponding environment during Gate training. This indicates that our Gate does not exhibit evident overfitting to the environments used during training.

\begin{figure}[t]
    \centering
    \includegraphics[width=0.99\linewidth]{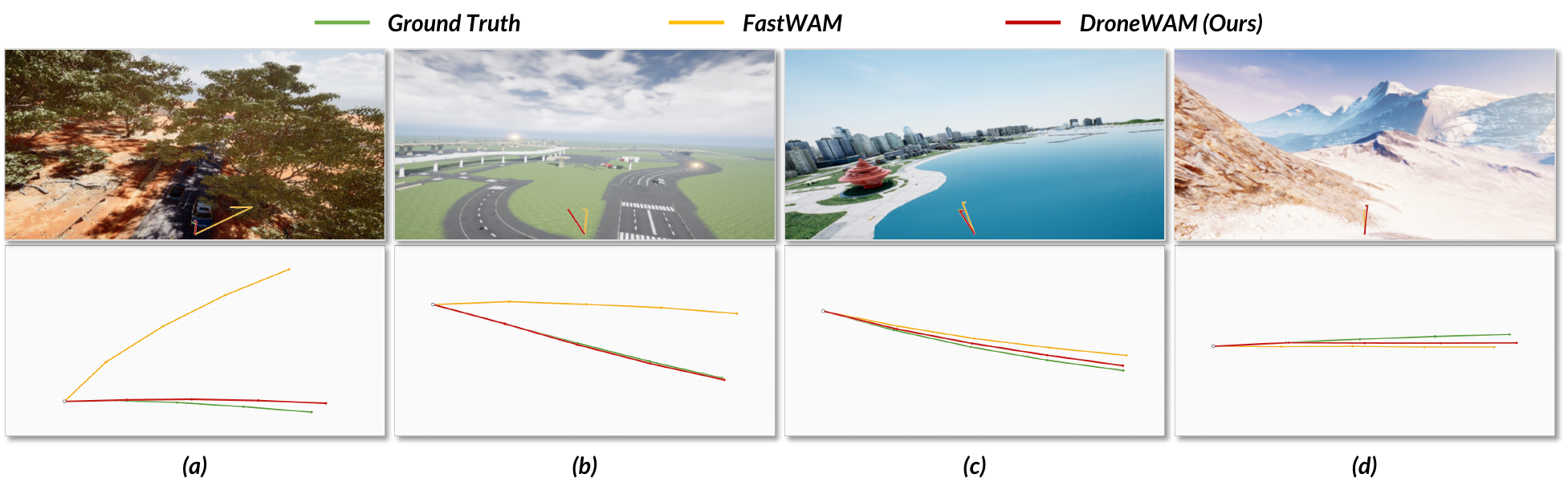}
    \caption{Qualitative comparison of predicted trajectories across diverse aerial scenes. Green, yellow, and red denote the ground-truth trajectory, FastWAM, and DroneWAM, respectively. DroneWAM consistently follows the reference trajectory more closely, while FastWAM exhibits larger directional and accumulated drift.}
    \label{vis}
\end{figure}

\paragraph{Qualitative Results.}
We qualitatively examine whether the improvements in trajectory metrics translate into more faithful motion prediction across different environments. Fig.~\ref{vis} compares the predicted trajectories of DroneWAM and FastWAM in four representative scenes. DroneWAM consistently remains closer to the ground-truth trajectory, preserving both the motion direction and trajectory shape more accurately. The difference is particularly evident when longer-term prediction amplifies small motion errors, where FastWAM exhibits substantial directional drift, while DroneWAM maintains a more stable trajectory. These examples complement the quantitative results, showing that the lower trajectory errors of DroneWAM correspond to more consistent motion prediction across diverse aerial environments.

\section{Related Work}
\label{sec:related_work}

\subsection{World Models and World-Action Models}

World models~\cite{azuma2026navwam,zhang2026world,hafner2019dream,ha2018world} learn predictive dynamics that allow an agent to reason about the consequences of its actions before execution. Early model-based agents~\cite{wang2026coevolving,chahe2026policy} typically learn compact latent dynamics and perform policy optimization through imagined trajectories, as exemplified by DreamerV3~\citep{hafner2025mastering}. Recent works increasingly move visual prediction into representation space. V-JEPA~2~\citep{assran2025v} and DINO-WM~\citep{zhou2024dino} predict action-conditioned future features and support planning without explicit RGB reconstruction, substantially reducing the complexity of modeling visual dynamics. World-Action Models further couple world prediction with action generation. DreamZero~\citep{ye2026world} jointly models future visual states and actions, while WAM-Nav~\citep{yang2026wam} uses asymmetric latent world-action modeling to balance visual foresight and action planning. These works establish predictive representations as a useful interface between perception and control. DroneWAM follows this direction while focusing on the inference cost of recurrent prediction. 

\subsection{Drone Navigation}

Drone navigation~\cite{yan2025sign,zheng2026onfly,sun2026autofly} has been studied under both visual goal conditioning and language-guided settings. Visual navigation methods use images, geometric goals, or predicted future observations to evaluate motion and select trajectories. Navigation World Models~\citep{bar2025navigation} demonstrate trajectory evaluation through controllable future prediction, while recent aerial extensions model long-range visual dynamics in three-dimensional environments~\citep{zhang2025aerial,zhu2026uncertainty}. Aerial vision-language navigation instead grounds natural-language instructions in egocentric observations. AerialVLN~\citep{liu2023aerialvln} established an outdoor UAV VLN setting, and subsequent benchmarks such as OpenUAV~\citep{wang2025towards}, CityNav~\citep{lee2025citynav}, and OpenFly~\citep{gao2026openfly} expand the scale, realism, and diversity of aerial navigation data. More recent approaches integrate predictive world modeling into aerial VLN. WorldVLN~\citep{zhao2026worldvln} performs autoregressive latent world-action prediction, while WorldFly~\citep{zheng2026worldfly} jointly generates future visual states and navigation actions. These developments bring explicit foresight to aerial navigation, although online prediction remains computationally demanding and most available navigation data still simplify aerial motion. Our work studies efficient world-action prediction together with higher-DoF aerial supervision.

\section{Conclusion}

We present DroneWAM, an efficient world-action model for drone visual navigation. DroneWAM combines JEPA-based latent prediction, compact visual representations, and adaptive rollout to reduce the cost of future imagination while maintaining strong navigation accuracy. We further introduce DroneNav-6D, a simulated dataset with synchronized visual observations and full 6-DoF aerial motion. Experiments demonstrate consistent gains in both prediction quality and inference efficiency, and show that adaptive rollout can allocate computation effectively across different scenes. We hope it can support compact and adaptive predictive modeling as a practical direction for autonomous drone navigation.

\section{Limitations}

DroneWAM's fixed token budget and bounded imagination horizon may require adjustment across scenes and computational budgets. Performance also depends on the information retained by its pretrained encoder and resampler. In simulation, closed-loop task success remains very low in complex scenes, leaving a substantial gap to reliable navigation. The current system is therefore not ready for real-world deployment. Practical use requires higher success rates, improved robustness, and validation on physical platforms. But, we believe compact world-action modeling with adaptive prediction remains a promising direction, with better representations, broader training coverage, and improved control integration offering a path toward reliable drone autonomy.

\bibliography{iclr2027_conference}
\bibliographystyle{iclr2027_conference}

\clearpage
\appendix

\section{Detailed Experimental Settings}
\label{app:experimental_settings}

\paragraph{Data and preprocessing.}
Training samples are constructed from 20-frame video clips recorded at 7\,Hz. The first four RGB frames and their associated motion states are used as observations, while the model predicts an eight-step action plan. Each motion state contains 3-D position, roll, pitch, yaw, and scalar speed. Actions contain ego-frame translation and rotation increments. Only observed states and previously executed actions are provided to the model; future states and actions are masked. Images are vertically cropped by 28 pixels at both boundaries, resized to $256\times512$, and normalized using ImageNet statistics.

\paragraph{Model configuration.}
The visual backbone is a frozen ViT-L with a patch size of 16, embedding width 1024, causal attention, and rotary positional embeddings. A pretrained TokenAE with four encoder layers, two decoder layers, and 16 attention heads reduces the number of visual tokens from 512 to 128. The TokenAE is frozen throughout training. The Predictor consists of 24 Transformer layers with width 1024, 16 attention heads, and an MLP expansion ratio of four. The Planner maintains a 256-D recurrent state and predicts an eight-step, 6-DoF action sequence. The goal-image representation is supplied to the Planner at every imagination depth.

\paragraph{Base training and adaptation.}
The base WAM is trained for 30 epochs using AdamW with bfloat16 precision, weight decay 0.04, and gradient clipping at 1.0. The learning rate is linearly increased from $10^{-6}$ to $7.5\times10^{-5}$ during the first two epochs and subsequently cosine-decayed to $10^{-6}$. We use a per-GPU batch size of 16 on eight A100-80GB GPUs, resulting in an effective batch size of 128. The visual encoder and TokenAE remain frozen, while the Predictor and action head are optimized. The action-regression loss is weighted by 0.35, and the autoregressive latent loss uses a temporal discount factor of 0.95. Translation dimensions receive larger action-loss weights than rotation dimensions.

After base training, the Predictor and Planner are jointly adapted for five epochs using rollouts of different imagination depths. The Predictor and Planner learning rates are $5\times10^{-8}$ and $2\times10^{-6}$, respectively. Losses are averaged over all evaluated rollout depths, while the visual encoder, TokenAE, and action head remain frozen.

\paragraph{Gate training.}
The marginal-value Gate contains five independently initialized ensemble members, each implemented as an MLP with hidden width 128. Gate inputs include the 256-D Planner state, the current action plan, changes in the Planner state and action plan from the previous depth, and a learned depth embedding. Each member is trained for 30 epochs using AdamW with a batch size of 512, learning rate $10^{-4}$, weight decay 0.04, and gradient clipping at 1.0. Bootstrap sampling is performed at the trajectory level, and 20\% of the training trajectories are reserved for threshold and uncertainty calibration. All experiments use random seed 239.

\section{Evaluation Metrics}
\label{app:evaluation_metrics}

\paragraph{Open-loop trajectory metrics.}
The predicted ego-frame actions are integrated from the final observed pose to obtain a global 6-DoF trajectory. Ego-frame translations are rotated into the global frame using the current yaw, while angular increments are accumulated and wrapped to $[-\pi,\pi]$. Let $\hat{\mathbf{p}}_t$ and $\mathbf{p}_t$ denote the predicted and ground-truth 3-D positions at step $t$, respectively, and let $H=8$ denote the prediction horizon. Absolute trajectory error (ATE) and relative pose error (RPE) are computed as
\begin{equation}
\mathrm{ATE}=\sqrt{\frac{1}{H+1}\sum_{t=0}^{H}\left\|\hat{\mathbf{p}}_t-\mathbf{p}_t\right\|_2^2},
\label{eq:ate}
\end{equation}
\begin{equation}
\mathrm{RPE}=\sqrt{\frac{1}{H}\sum_{t=1}^{H}\left\|(\hat{\mathbf{p}}_t-\hat{\mathbf{p}}_{t-1})-(\mathbf{p}_t-\mathbf{p}_{t-1})\right\|_2^2}.
\label{eq:rpe}
\end{equation}
Here, RPE measures the error between consecutive translation increments; rotational accuracy is evaluated separately. To account for different trajectory scales, both errors are normalized by the ground-truth path length:
\begin{equation}
L=\sum_{t=1}^{H}\left\|\mathbf{p}_t-\mathbf{p}_{t-1}\right\|_2,\qquad \mathrm{rel.ATE}=\frac{\mathrm{ATE}}{L},\qquad \mathrm{rel.RPE}=\frac{\mathrm{RPE}}{L}.
\label{eq:relative_metrics}
\end{equation}
The normalization is performed independently for each trajectory before averaging over the evaluation set. Lower values indicate better trajectory prediction.

\paragraph{Attitude and prediction metrics.}
Let $\hat{\boldsymbol{\theta}}_t$ and $\boldsymbol{\theta}_t$ denote the predicted and ground-truth roll, pitch, and yaw. Attitude error is computed as the mean absolute wrapped angular difference:
\begin{equation}
E_{\mathrm{att}}=\frac{1}{3(H+1)}\sum_{t=0}^{H}\sum_{k=1}^{3}\left|\operatorname{wrap}\!\left(\hat{\theta}_{t,k}-\theta_{t,k}\right)\right|.
\label{eq:attitude_error}
\end{equation}
We additionally report the unweighted RMSE over all six action dimensions and the mean absolute error between predicted and target future latent tokens:
\begin{equation}
E_{\mathrm{act}}=\sqrt{\frac{1}{6H}\sum_{t=1}^{H}\left\|\hat{\mathbf{a}}_t-\mathbf{a}_t\right\|_2^2},\qquad E_{\mathrm{latent}}=\frac{1}{N_z}\left\|\hat{\mathbf{z}}-\mathbf{z}\right\|_1,
\label{eq:prediction_metrics}
\end{equation}
where $N_z$ is the number of evaluated latent elements. Action and latent errors are diagnostic metrics, while rel.ATE and rel.RPE are used as the primary trajectory-quality measures.

\paragraph{Closed-loop task metrics.}
During closed-loop evaluation, only the first action of each predicted plan is executed before obtaining a new observation and replanning. The trajectory metrics above are computed from the complete executed trajectory and its time-aligned reference trajectory without post-hoc spatial or temporal alignment. Task progress is computed by projecting the terminal position onto the nearest segment of the full reference path and normalizing the resulting arc length by the total reference length. Terminal position and attitude errors are defined as
\begin{equation}
E_{\mathrm{goal}}^{\mathrm{pos}}=\left\|\mathbf{p}_{T}-\mathbf{p}_{\mathrm{goal}}\right\|_2,\qquad E_{\mathrm{goal}}^{\mathrm{att}}=\frac{1}{3}\sum_{k=1}^{3}\left|\operatorname{wrap}\!\left(\theta_{T,k}-\theta_{\mathrm{goal},k}\right)\right|.
\label{eq:terminal_errors}
\end{equation}
An episode is considered successful only when it completes the prescribed control horizon without collision and terminates within 2\,m position error and 0.35\,rad attitude error. Success and collision rates are computed as the corresponding fractions over all evaluation episodes.

\paragraph{Computational efficiency.}
Inference latency is measured with batch size one using synchronized GPU wall-clock time, excluding data loading and image preprocessing. We report mean, median, and 95th-percentile latency. Throughput and the Predictor-step reduction of the adaptive model are calculated as
\begin{equation}
\mathrm{FPS}=\frac{1000}{\overline{t}_{\mathrm{infer}}},\qquad \mathrm{Reduction}=1-\frac{\overline{d}}{H},
\label{eq:efficiency}
\end{equation}
where $\overline{t}_{\mathrm{infer}}$ is the mean inference latency in milliseconds and $\overline{d}$ is the mean selected rollout depth. Trajectory metrics are averaged equally across episodes, whereas latency and rollout depth are averaged over all executed control steps.

\section{DroneNav-6D Dataset Analysis}
\label{app:dataset_analysis}

We analyze the size, trajectory characteristics, six-dimensional motion, and visual appearance of DroneNav-6D. The statistics cover five simulation worlds: CITYBIM, NYC, RuralAustralia, SnappyRoads, and Snowy Mountain. The following analyses use recorded trajectory clips, pose increments, and sampled RGB observations.

\subsection{Dataset Size and Composition}
\label{app:dataset_size}

We first quantify the amount of recorded flight data and its distribution across environments to characterize the supervision available for learning future prediction.

\begin{figure}[h]
  \centering
  \includegraphics[width=\linewidth]{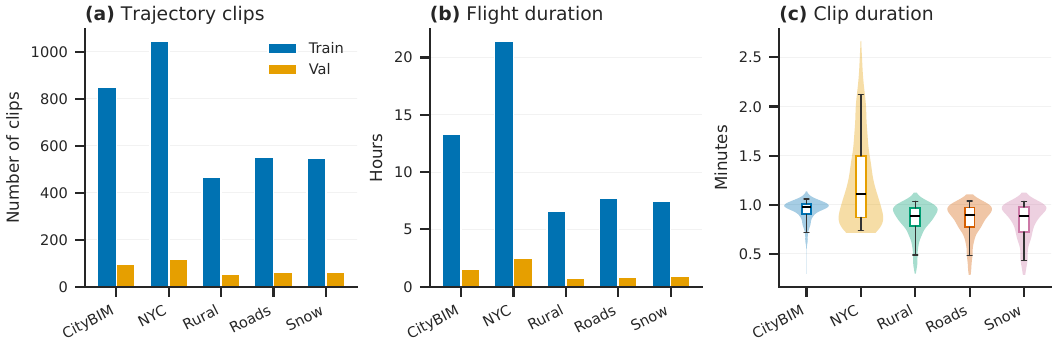}
  \caption{Dataset composition across the five worlds. The panels show (a) training and validation clip counts, (b) recorded flight duration, and (c) the distribution of individual clip durations. CityBIM, Rural, Roads, and Snow denote CITYBIM, RuralAustralia, SnappyRoads, and Snowy Mountain, respectively. These abbreviations are used throughout the dataset figures.}
  \label{fig:dataset_composition}
\end{figure}

Table~\ref{tab:dataset_size} summarizes 3,837 trajectory clips containing 1,592,170 frames, 63.03 hours of flight, and 2,004.11\,km of traveled distance. The training and validation sets contain 3,453 and 384 clips, respectively, with distinct clip identifiers and all five worlds represented in both sets. The median recording rate is approximately 7.00\,Hz. The clips provide 1,519,267 overlapping windows of 20 consecutive frames, each contained within a single clip. Figure~\ref{fig:dataset_composition} shows the contribution of each world. NYC contributes 603,705 frames, or 37.92\% of the corpus, followed by CITYBIM with 373,702 frames, or 23.47\%. The other three worlds each contribute between 11.64\% and 13.66\% of the frames. The overall median clip duration is 57.12\,s, with an interquartile range of 49.42--61.43\,s.

These statistics establish a corpus of continuous flight sequences with temporal context for multi-step prediction and coverage of all five worlds in the training and validation data.

\begin{table}[h]
  \centering
  \caption{Size of DroneNav-6D. Window counts enumerate overlapping sequences of 20 consecutive frames within individual clips. Flight duration and distance are summed over clips.}
  \label{tab:dataset_size}
  \begin{tabular}{@{}l|rrrrr@{}}
    \toprule
    Split & Clips & Frames & Hours & Distance (km) & 20-frame windows \\
    \midrule
    Training & 3,453 & 1,427,016 & 56.49 & 1,795.42 & 1,361,409 \\
    Validation & 384 & 165,154 & 6.54 & 208.69 & 157,858 \\
    \midrule
    Total & 3,837 & 1,592,170 & 63.03 & 2,004.11 & 1,519,267 \\
    \bottomrule
  \end{tabular}
\end{table}

\subsection{Six-Dimensional Motion Distributions}
\label{app:motion_distributions}

We examine the distributions of pose increments to characterize the translation and attitude changes represented in each of the six motion components.

\begin{figure}[h]
  \centering
  \includegraphics[width=\linewidth]{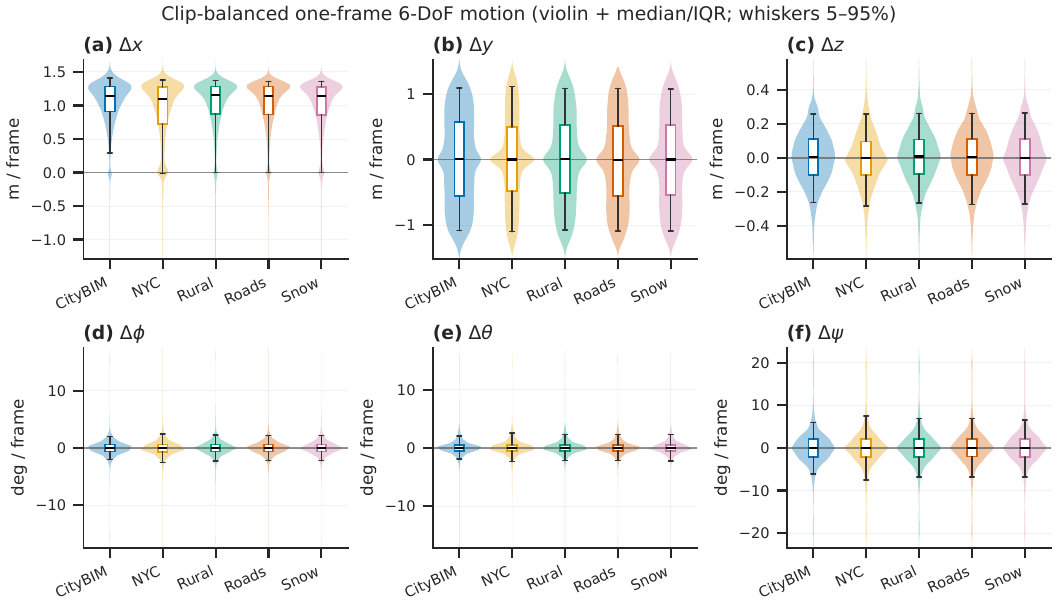}
  \caption{Distributions of six-dimensional pose increments between consecutive frames, shown separately for each world using clip-balanced samples. The top row shows translation in meters per frame, and the bottom row shows roll, pitch, and yaw changes in degrees per frame. Violin plots depict the distributions; boxes indicate the interquartile range, central lines mark the medians, and whiskers span the 5th--95th percentiles.}
  \label{fig:motion_distributions}
\end{figure}

For this analysis, $\Delta x$ and $\Delta y$ express translation in horizontal axes aligned with the current yaw, and $\Delta z$ follows the NED vertical axis. The angular components $\Delta\phi$, $\Delta\theta$, and $\Delta\psi$ are wrapped differences in roll, pitch, and yaw. Figure~\ref{fig:motion_distributions} presents the per-world distributions using clip-balanced samples. Forward translation is concentrated at positive values, while lateral and vertical translation include changes in both directions. All five worlds also exhibit positive and negative roll, pitch, and yaw increments. Table~\ref{tab:pose_increments} provides the corresponding dataset summary: the median forward increment is 1.138\,m per frame, and the 10th--90th percentile intervals are $[-1.395, 1.392]^{\circ}$ per frame for roll, $[-1.252, 1.376]^{\circ}$ per frame for pitch, and $[-4.406, 4.376]^{\circ}$ per frame for yaw. Yaw exhibits a wider central range than roll and pitch.

These distributions show that the recorded motion combines forward flight, lateral and vertical adjustments, and variation in all three attitude components. The resulting pose increments provide supervision for learning future motion across all six degrees of freedom.

\begin{table}[h]
  \centering
  \caption{Distribution of pose increments between consecutive frames. Translation uses yaw-aligned horizontal axes and the NED vertical axis; angular increments are wrapped Euler-angle differences. P10 and P90 denote the 10th and 90th percentiles. Values are rounded to three decimal places.}
  \label{tab:pose_increments}
  \setlength{\tabcolsep}{10pt}
  \begin{tabular}{@{}ll|rrr@{}}
    \toprule
    Component & Unit & P10 & Median & P90 \\
    \midrule
    $\Delta x$ & m/frame & 0.408 & 1.138 & 1.346 \\
    $\Delta y$ & m/frame & $-0.943$ & 0.000 & 0.945 \\
    $\Delta z$ & m/frame & $-0.199$ & 0.000 & 0.193 \\
    $\Delta\phi$ (roll) & deg/frame & $-1.395$ & 0.000 & 1.392 \\
    $\Delta\theta$ (pitch) & deg/frame & $-1.252$ & 0.011 & 1.376 \\
    $\Delta\psi$ (yaw) & deg/frame & $-4.406$ & 0.000 & 4.376 \\
    \bottomrule
  \end{tabular}
\end{table}

\subsection{Trajectory Length and Motion}
\label{app:trajectory_characteristics}

We compare clip duration, traveled distance, vertical travel, and flight speed to understand how trajectory characteristics vary across environments.

\begin{figure}[h]
  \centering
  \includegraphics[width=\linewidth]{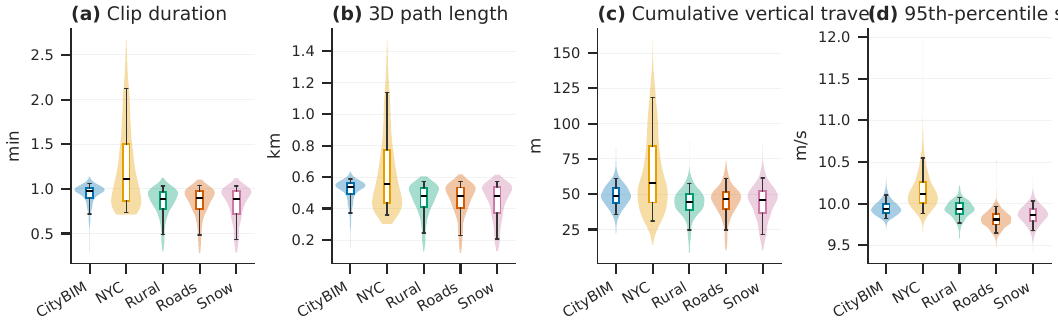}
  \caption{Trajectory characteristics across worlds. Violin and box plots summarize (a) clip duration, (b) three-dimensional path length, (c) cumulative vertical travel, and (d) the 95th-percentile linear speed within each clip.}
  \label{fig:trajectory_characteristics}
\end{figure}

Figure~\ref{fig:trajectory_characteristics} summarizes clip duration, three-dimensional path length, cumulative vertical travel, and the 95th-percentile linear speed within each clip. NYC exhibits wider distributions of duration, path length, and vertical travel. Its median clip duration is 66.57\,s, compared with 53.14--58.44\,s across the other worlds (Table~\ref{tab:world_statistics}). The path-length medians are approximately 0.5\,km, with a longer upper tail in NYC. Cumulative vertical travel also extends to larger values in NYC. Across worlds, the median flight speeds lie in the narrow range of 9.298--9.357\,m/s, while typical clip-level 95th-percentile speeds are close to 10\,m/s.

The dataset therefore combines similar typical flight speeds with variation in sequence duration, traveled distance, and vertical travel. These quantities describe complementary aspects of the motion represented in the collected trajectories.

\begin{table}[h]
  \centering
  \caption{Per-world data and motion statistics. Duration is the median clip duration; speed is the per-world median in the dataset statistics.}
  \label{tab:world_statistics}
  \setlength{\tabcolsep}{7pt}
  \begin{tabular}{@{}l|rrrr@{}}
    \toprule
    World & Clips & Frames & Duration (s) & Speed (m/s) \\
    \midrule
    CITYBIM & 943 & 373,702 & 58.44 & 9.357 \\
    NYC & 1,159 & 603,705 & 66.57 & 9.298 \\
    RuralAustralia & 518 & 185,346 & 53.14 & 9.319 \\
    SnappyRoads & 611 & 217,483 & 53.58 & 9.329 \\
    Snowy Mountain & 606 & 211,934 & 53.16 & 9.325 \\
    \bottomrule
  \end{tabular}
\end{table}

\subsection{Motion over Longer Horizons}
\label{app:horizon_motion}

We examine how translational and rotational changes grow with the prediction horizon to characterize the motion encountered during multi-step future prediction.

\begin{figure}[h]
  \centering
  \includegraphics[width=\linewidth]{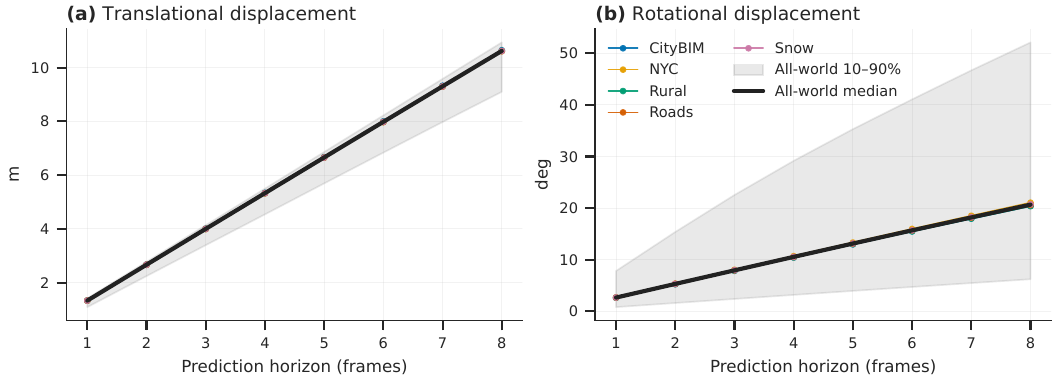}
  \caption{Motion over prediction horizons of one to eight frame intervals. The panels show translational displacement and quaternion-based geodesic rotation. Colored curves are per-world medians; the black curve is the pooled median, and the shaded region spans the pooled 10th--90th percentiles.}
  \label{fig:horizon_motion}
\end{figure}

Figure~\ref{fig:horizon_motion} summarizes translational displacement and quaternion-based geodesic rotation over horizons of one to eight frame intervals. The median translational displacement increases approximately linearly from 1.3\,m at one interval to 10.6\,m at eight intervals. The median rotational displacement grows from approximately $2.7^{\circ}$ to $20.7^{\circ}$. The per-world median curves remain close throughout this range. The pooled 10th--90th percentile interval for rotation broadens substantially, reaching approximately $6^{\circ}$--$52^{\circ}$ at eight intervals. At the median recording rate of 7.00\,Hz, the longest horizon corresponds to approximately 1.14\,s.

These observations show that longer prediction horizons cover larger position and attitude changes, together with a broad range of rotational changes across samples. They motivate evaluating multi-step prediction across both the horizon length and the amount of viewpoint change.

\subsection{Translation and Rotation}
\label{app:translation_rotation}

We examine correlations between translation and rotation at the single-frame and clip levels to characterize how these aspects of flight vary together.

\begin{figure}[h]
  \centering
  \includegraphics[width=\linewidth]{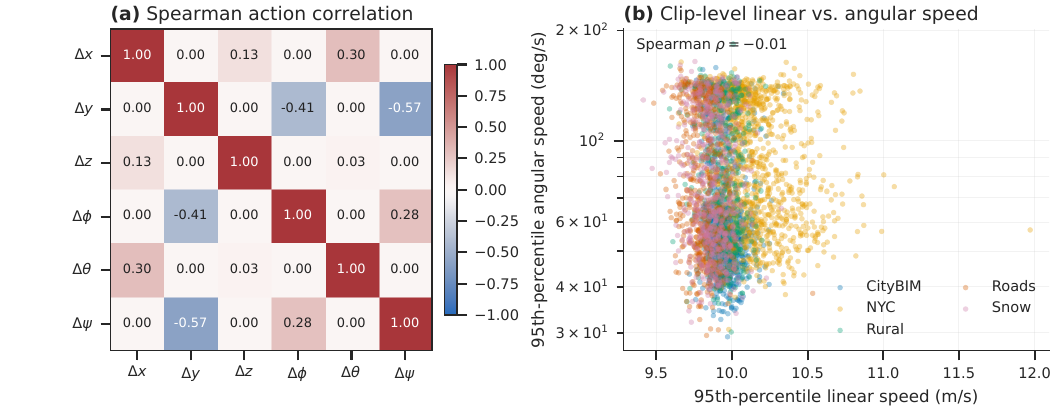}
  \caption{Translation and rotation in the recorded motion. (a) Spearman correlations among the six pose-increment components. (b) Each point shows the 95th-percentile linear and angular speeds of one clip, colored by world. The vertical axis uses a logarithmic scale.}
  \label{fig:motion_coupling}
\end{figure}

Figure~\ref{fig:motion_coupling}(a) shows associations between lateral translation and yaw ($\rho=-0.573$), lateral translation and roll ($\rho=-0.413$), and forward translation and pitch ($\rho=0.297$), using the pose-increment definitions in Section~\ref{app:motion_distributions}. At the clip level, the 95th-percentile linear and angular speeds have a Spearman correlation of approximately $-0.01$ (Figure~\ref{fig:motion_coupling}(b)). Clips with similar linear speeds span a broad range of angular speeds.

The single-frame statistics identify associations among translational and angular components, and the clip-level results show variation in rotational motion at similar flight speeds. Together, they motivate describing trajectories with both translation and attitude information and reporting angular motion alongside linear speed.

\subsection{Visual Appearance across Environments}
\label{app:visual_appearance}

We examine visual features from sampled RGB observations to characterize appearance variation across worlds and the local feature neighborhoods covered by the training and validation samples.

\begin{figure}[h]
  \centering
  \includegraphics[width=\linewidth]{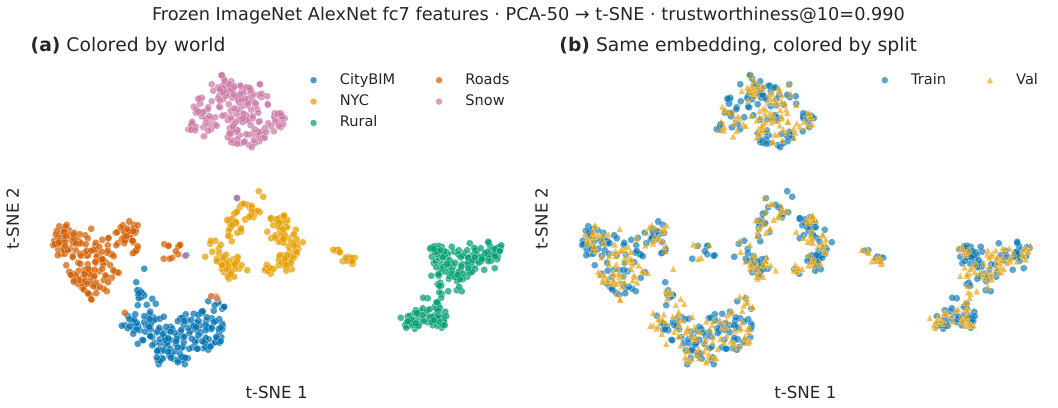}
  \caption{Visual features of 1,000 sampled RGB observations. Each world contributes 100 training and 100 validation images. Frozen ImageNet-pretrained AlexNet fc7 features are projected with PCA to 50 dimensions and then with t-SNE to two dimensions. The same coordinates are colored by (a) world and (b) split. The reported trustworthiness at 10 neighbors is 0.990; the visualization characterizes local feature neighborhoods.}
  \label{fig:visual_appearance}
\end{figure}

Figure~\ref{fig:visual_appearance} uses 1,000 images, with 100 training images and 100 validation images sampled from each world. Frozen ImageNet-pretrained AlexNet fc7 features are reduced to 50 dimensions using PCA and then projected into two dimensions using t-SNE. The reported trustworthiness at 10 neighbors is 0.990. When colored by world, samples form groups associated with the individual environments, showing world-specific appearance patterns in this feature representation. Recoloring the same embedding by split shows training and validation samples throughout the local groups.

The visualization provides qualitative evidence of appearance variation across the five worlds and shared local visual coverage between the sampled training and validation images within these environments.

\section{More Data Visualization}
\label{app:data_vis}

\begin{figure}[h]
  \centering
  \includegraphics[width=0.7\linewidth]{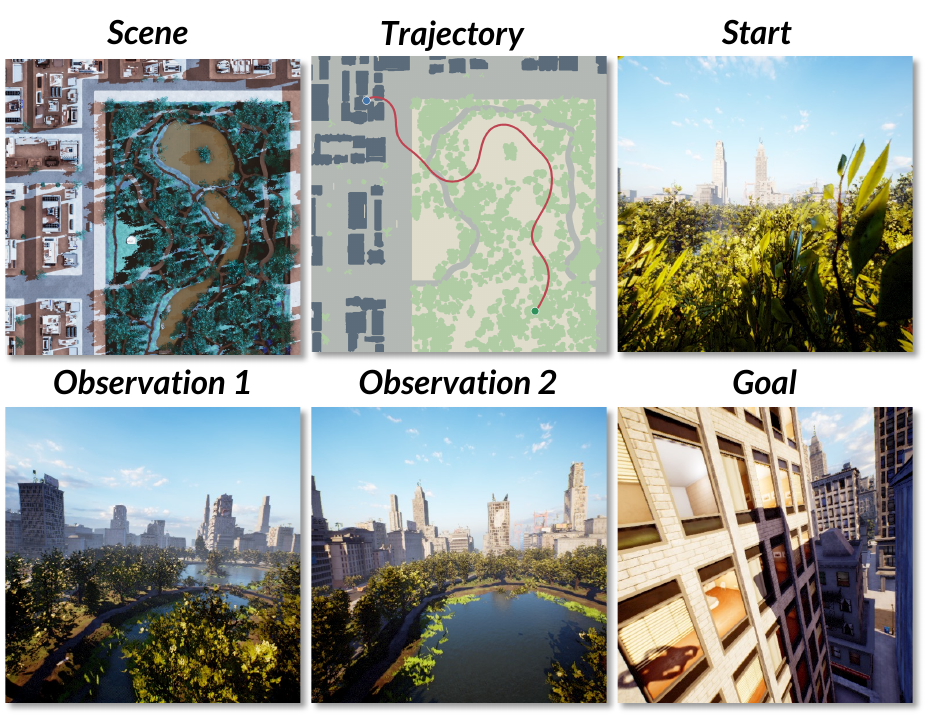}
  \caption{DroneNav-6D sample 1.}
  \label{fig:data1}
\end{figure}

\begin{figure}[h]
  \centering
  \includegraphics[width=0.7\linewidth]{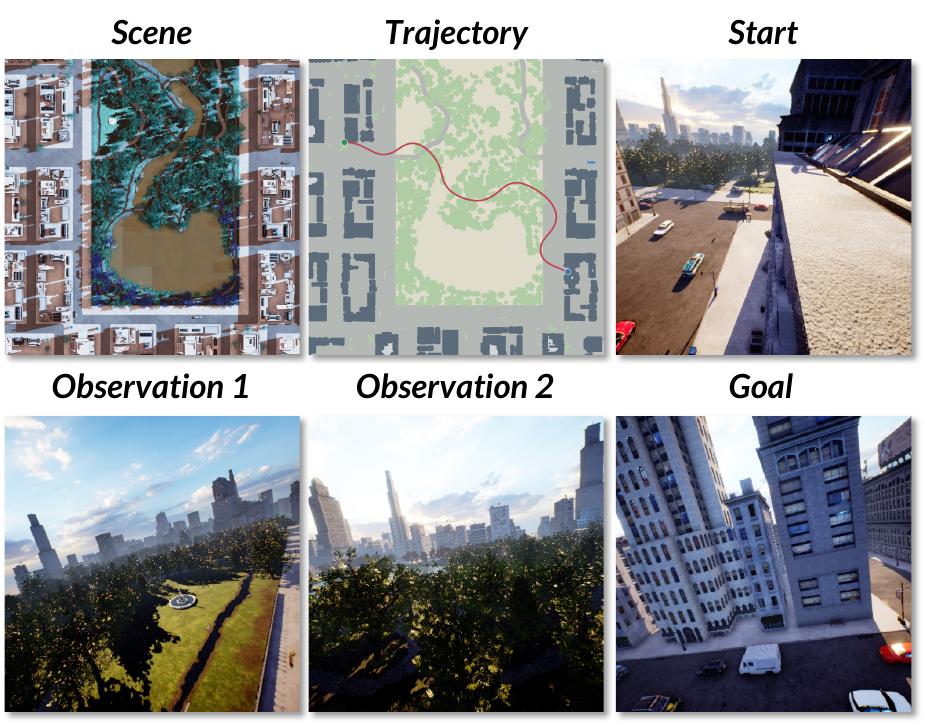}
  \caption{DroneNav-6D sample 2.}
  \label{fig:data2}
\end{figure}

Figures~\ref{fig:data1} and ~\ref{fig:data2} present two representative goal-conditioned navigation samples from DroneNav-6D, together with the global scene layout, flight trajectory, start observation, intermediate observations, and goal image. In the first example, the drone starts above a vegetation-dominated region and follows a long, curved trajectory across the central park before approaching a building-dominated goal location. The intermediate observations gradually reveal the water body and surrounding skyline, while the goal view differs substantially from the initial observation in both scene appearance and camera viewpoint. In the second example, the drone starts close to urban structures, transitions through an open park region, and eventually reaches a dense building area. Although the trajectory is comparatively smoother, the visual observations still undergo substantial changes in scene composition and viewpoint throughout the flight.

These examples illustrate that image-goal navigation in DroneNav-6D cannot rely on direct visual matching between the current observation and the goal image. Instead, the model needs to capture the underlying motion and scene evolution over extended 6-DoF trajectories in order to predict actions that consistently progress toward the target. More broadly, DroneNav-6D intentionally contains relatively long navigation paths with substantial visual changes between the start and goal observations, making long-horizon navigation from visual observations particularly challenging. This difficulty is also reflected in the relatively limited closed-loop task completion of current models. Nevertheless, we believe that continued advances in world models, especially in long-horizon prediction and action-conditioned imagination, will progressively enable more reliable long-range navigation from visual observations alone.


\clearpage

\section{Additional Experimental Results}
\label{app:additional-experiments}

\paragraph{Closed-loop Results on Easy Routes.} 
The challenging routes in our original closed-loop benchmark require long-horizon visual localization and repeated replanning, under which none of the evaluated methods successfully completes an episode. To further examine whether the models can nevertheless support effective closed-loop navigation under a more tractable setting, we construct an additional easy-route test set containing 50 trajectories. These routes consist primarily of straight flight segments and mild turns, reducing the difficulty of long-horizon visual localization while retaining the same closed-loop perception--prediction--action evaluation protocol. As shown in Table~\ref{tab:straight_closed_loop_global}, DroneWAM achieves an 88\% success rate, outperforming NWM and FastWAM by 14 and 16 percentage points, respectively. It also substantially reduces trajectory error, achieving a rel.ATE of 0.15 and a rel.RPE of 0.008, compared with 0.31/0.012 for NWM and 0.38/0.014 for FastWAM. Meanwhile, DroneWAM reaches a task progress of 0.95, indicating that its predicted actions remain well aligned with the reference route throughout closed-loop execution. Importantly, these gains are obtained with an inference latency of 460\,ms, lower than FastWAM's 700\,ms and substantially lower than NWM. These results verify that DroneWAM can support successful closed-loop visual navigation when route complexity is moderate. Together with the results on the full benchmark, they suggest that the remaining failures are primarily associated with challenging long-horizon visual localization and accumulated control errors, rather than an inability of the model to operate in a closed-loop manner. 

\begin{table}[h]
\centering
\caption{Global visual closed-loop results on easy routes.}
\label{tab:straight_closed_loop_global}
\begin{tabular}{lccccc}
\toprule
Method & Success $\uparrow$ &
rel.ATE $\downarrow$ & rel.RPE $\downarrow$ &
Task Progress $\uparrow$ & Latency (ms) $\downarrow$\\
\midrule
NWM & 74\%& 0.31& 0.012& 0.79& $1.3\times10^5$\\
FastWAM & 72\%& 0.38& 0.014& 0.76& 700ms\\
DroneWAM & 88\%& 0.15& 0.008& 0.95& 460ms\\
\bottomrule
\end{tabular}
\end{table}

\paragraph{Quantitative Case Studies of Different Rollout Depths.}
To better understand how rollout depth affects individual predictions, we examine four representative held-out examples. For each example, we report the prediction errors at depth 1, the depth selected by the Gate, and depth 8.
As shown in Table~\ref{tab:appendix-rollout-cases}, the effect of additional rollout varies across samples. In Case 1, increasing the rollout depth substantially improves the prediction, reducing rel.ATE from 0.1879 at depth 1 to 0.1083 at the Gate-selected depth 4, with only a small further improvement at depth 8. Case 2 shows the opposite trend: the prediction is already accurate at depth 1, while deeper rollout slightly increases both rel.ATE and rel.RPE. In Case 3, the Gate selects depth 2, where the lowest error is obtained; continuing to depth 8 instead increases rel.ATE from 0.0423 to 0.0514. Case 4 illustrates a failure case, where the Gate stops at depth 1 although deeper rollout would lead to a substantially lower error. These examples further show that the preferred rollout depth is sample-dependent. Some observations benefit from additional prediction, while others are already sufficiently resolved after only a few steps. They also reveal a remaining limitation of the learned Gate: inaccurate early stopping can occasionally prevent the model from benefiting from deeper rollout.

\begin{table*}[h]
\centering
\caption{Case studies at different rollout depths. Each row evaluates the same sample at depth 1, the Gate-selected depth, and depth 8.}
\label{tab:appendix-rollout-cases}
\resizebox{\textwidth}{!}{%
\begin{tabular}{l|c|cccccc}
\toprule
Case & Gate depth & D1 rel.ATE & Gate rel.ATE & D8 rel.ATE & D1 rel.RPE & Gate rel.RPE & D8 rel.RPE \\
\midrule
Case 1 & 4 & 0.1879 & 0.1083 & \textbf{0.1064} & 0.0627 & 0.0366 & \textbf{0.0322} \\
Case 2 & 1 & \textbf{0.0142} & \textbf{0.0142} & 0.0192 & \textbf{0.0064} & \textbf{0.0064} & 0.0073 \\
Case 3 & 2 & 0.0543 & \textbf{0.0423} & 0.0514 & 0.0164 & \textbf{0.0127} & 0.0150 \\
Case 4 & 1 & 0.2160 & 0.2160 & \textbf{0.1285} & 0.0924 & 0.0924 & \textbf{0.0668} \\
\bottomrule
\end{tabular}}
\end{table*}


\paragraph{Does Adaptive Rollout Transfer Beyond Aerial Motion?}
The main paper reports only normalized RMSE, gripper accuracy, and latency on LIBERO. To verify that the conclusion is consistent across action metrics, we evaluate normalized RMSE and L1 over all seven action dimensions, raw-unit end-effector RMSE and L1 over the first six dimensions, gripper-state accuracy, mean rollout depth, and synchronized batch-one latency. The episode-level split contains 173 held-out trajectories, and every method predicts the same 32-step action chunk from the same deterministic center window.
As shown in Table~\ref{tab:appendix-libero-full}, adaptive rollout consistently improves the accuracy--efficiency trade-off. Compared with Base WAM, our method reduces normalized RMSE from 0.3636 to 0.2652 and latency from 957.1\,ms to 411.9\,ms, corresponding to reductions of 27.1\% and 57.0\%, respectively. Compared with Fixed D8, it reduces latency by 53.8\%, while further improving normalized RMSE from 0.2685 to 0.2652 and gripper accuracy from 89.83\% to 90.55\%, an increase of 0.72 percentage points. The same trend is observed for normalized L1 and both end-effector metrics.
Importantly, the adaptive model achieves the best action-prediction accuracy across all reported metrics with an average rollout depth of only 3.56. It also outperforms Fixed D4 in both prediction quality and latency, despite using fewer rollout steps on average. These results indicate that the gain does not arise from simply truncating imagination, but from allocating predictive computation according to the input. Overall, adaptive rollout transfers effectively beyond aerial navigation to robotic manipulation, reducing computation without sacrificing action-prediction quality.

\begin{table*}[h]
\centering
\caption{Full open-loop action-prediction results.}
\label{tab:appendix-libero-full}
\resizebox{\textwidth}{!}{%
\begin{tabular}{l|ccccccc}
\toprule
Method & Mean depth & Norm. RMSE $\downarrow$ & Norm. L1 $\downarrow$ & EEF RMSE $\downarrow$ & EEF L1 $\downarrow$ & Gripper acc. $\uparrow$ & Latency (ms) $\downarrow$ \\
\midrule
Base WAM & -- & 0.3636 & 0.2466 & 0.2647 & 0.1631 & 85.42\% & 957.1 \\
Fixed D1 & 1.00 & 0.2735 & 0.1874 & 0.1932 & 0.1232 & 89.79\% & \textbf{129.7} \\
Fixed D2 & 2.00 & 0.2680 & 0.1844 & 0.1906 & 0.1218 & 90.21\% & 237.7 \\
Fixed D4 & 4.00 & 0.2667 & 0.1839 & 0.1890 & 0.1212 & 90.35\% & 451.3 \\
Fixed D8 & 8.00 & 0.2685 & 0.1848 & 0.1905 & 0.1223 & 89.83\% & 890.8 \\
Ours & 3.56 & \textbf{0.2652} & \textbf{0.1825} & \textbf{0.1881} & \textbf{0.1206} & \textbf{90.55\%} & 411.9 \\
\bottomrule
\end{tabular}}
\end{table*}

\paragraph{Does the gain hold across all LIBERO suites?}
We next disaggregate normalized action RMSE by suite to test whether the aggregate result is driven by one task family. Table~\ref{tab:appendix-libero-suite} shows that Adaptive WAM improves over Fixed D8 on Spatial, Object, Goal, and Long, with absolute reductions of 0.0046, 0.0026, 0.0037, and 0.0021, respectively. The selected depth varies from 3.09 on Goal to 4.02 on Object, demonstrating that the Gate changes its compute allocation across task distributions. Goal and Long remain substantially harder in absolute error, but the adaptive advantage is consistent across all four suites. This result supports method-level transfer rather than a gain restricted to a single manipulation category.

\begin{table}[h]
\centering
\caption{Normalized action RMSE by LIBERO suite.}
\label{tab:appendix-libero-suite}
\begin{tabular}{l|ccccc}
\toprule
Suite & Episodes & Base WAM $\downarrow$ & Fixed D8 $\downarrow$ & Ours $\downarrow$ & Mean depth \\
\midrule
Spatial & 44 & 0.3479 & 0.2321 & \textbf{0.2275} & 3.43 \\
Object & 46 & 0.2609 & 0.1857 & \textbf{0.1831} & 4.02 \\
Goal & 44 & 0.4461 & 0.3478 & \textbf{0.3441} & 3.09 \\
Long & 39 & 0.4094 & 0.3177 & \textbf{0.3156} & 3.69 \\
\bottomrule
\end{tabular}
\end{table}


\paragraph{How to Improve Closed-Loop Long-Horizon Manipulation?}
Our previous LIBERO experiments mainly evaluate open-loop action prediction, where lower action error does not necessarily translate into successful closed-loop execution. In long-horizon manipulation, small prediction errors can accumulate over repeated execution and replanning, while visually similar observations may correspond to different tasks depending on the language instruction. This makes explicit task conditioning particularly important for maintaining goal-consistent behavior over long horizons.

To improve task disambiguation, we inject the language instruction into both latent imagination and action planning. Specifically, a task-goal encoder maps the instruction to a compact task embedding, which is supplied to a task-conditioned future adapter to modulate the autoregressively predicted latent tokens. The same task embedding is also provided to the iterative Planner, allowing action decoding to depend jointly on the observed visual context, imagined future, robot state, executed-action history, and task specification. We retain the causal closed-loop interface. The improved policy uses four causal dual-view RGB observations, a fixed rollout depth of \(D=2\), and a 32-step predicted action chunk, of which the first 16 controls are executed before replanning from fresh observations. The visual encoder, TokenAE resampler, and pretrained WAM backbone remain frozen, while the task-goal encoder, task-conditioned future adapter, Plan Adapter, and action generator are optimized on the LIBERO training demonstrations.

As shown in Table~\ref{tab:appendix-libero-heldout}, explicit task conditioning substantially improves closed-loop manipulation across all four LIBERO suites. The average success rate increases from 45.23\% to 59.94\%, corresponding to an absolute gain of 14.71 percentage points. The largest improvement is observed on Object, where success rises from 64.32\% to 90.22\%, followed by Goal with a gain of 15.95 points and Spatial with a gain of 10.63 points. These improvements are consistent with the role of language conditioning in disambiguating task intent and keeping both imagined futures and action plans aligned with the specified goal.

The Long suite remains the most challenging setting, although its success rate improves from 10.08\% to 16.44\%. This suggests that task conditioning alleviates goal ambiguity but does not fully resolve the accumulated prediction and control errors that arise over extended closed-loop horizons. More broadly, the results indicate that accurate open-loop action prediction alone is insufficient for long-horizon manipulation: effective closed-loop behavior also requires the world model to preserve task-relevant information throughout imagination and planning. We expect further improvements in long-horizon world modeling, action-conditioned prediction, and closed-loop replanning to progressively narrow this gap.

\begin{table*}[h]
\centering
\caption{Closed-loop success rates (\%) on LIBERO.}
\label{tab:appendix-libero-heldout}
\begin{tabular}{l|rrrrr}
\toprule
Method & Spatial & Object & Goal & Long & Avg. \\
\midrule
DroneWAM & 52.26 & 64.32 & 54.27 & 10.08 & 45.23 \\
Improved-DroneWAM & \textbf{62.89} & \textbf{90.22} & \textbf{70.22} & \textbf{16.44} & \textbf{59.94} \\
\bottomrule
\end{tabular}
\end{table*}

\clearpage
\section{Algorithm}

\begin{algorithm}[h]
\caption{Training DroneWAM}
\label{alg:dronewam_training}
\footnotesize
\begin{algorithmic}[1]
\Require Training set $\mathcal{D}_{\mathrm{tr}}$, calibration set $\mathcal{D}_{\mathrm{cal}}$, frozen encoder $E$, maximum rollout depth $K_{\max}$, action horizon $H$
\Ensure Resampler $R_{\rho}$, Predictor $F_{\theta}$, Planner $P_{\psi}$, Gate $p_{\omega}$, and thresholds $\{\eta_k\}_{k=K_{\min}}^{K_{\max}-1}$

\Statex \textbf{Stage I: Resampler pretraining}
\For{each minibatch of video clips from $\mathcal{D}_{\mathrm{tr}}$}
    \State Extract dense visual tokens $\mathbf{x}_{1:T} \gets E(\mathbf{o}_{1:T})$
    \State Compress tokens $\mathbf{z}_{1:T} \gets R_{\rho}(\mathbf{x}_{1:T})$
    \State Reconstruct $\widetilde{\mathbf{x}}_{1:T} \gets \mathrm{Decoder}(\mathbf{z}_{1:T})$
    \State Update $R_{\rho}$ and Decoder using feature-reconstruction and temporal-consistency losses
\EndFor
\State Discard Decoder and freeze $E$ and $R_{\rho}$

\Statex \textbf{Stage II: Predictor--Planner training}
\For{each $(\mathbf{o}_{\leq t},\mathbf{s}_{\leq t},\mathbf{a}_{\leq t},g)$ from $\mathcal{D}_{\mathrm{tr}}$}
    \State $\mathbf{z}_{\leq t} \gets R_{\rho}(E(\mathbf{o}_{\leq t}))$ and $\mathbf{z}_{g} \gets R_{\rho}(E(g))$
    \State Form the causal context $C_t \gets \{\mathbf{z}_{\leq t},\mathbf{s}_{\leq t},\mathbf{a}_{\leq t}\}$
    \For{$k=1,\ldots,K_{\max}$}
        \State $\widehat{\mathbf{z}}_{t+k} \gets F_{\theta}(C_t,\mathbf{z}_g,\widehat{\mathbf{z}}_{t+1:t+k-1})$
        \State $\widehat{\mathbf{A}}_{t}^{(k)} \gets P_{\psi}(\widehat{\mathbf{z}}_{t+1:t+k},\mathbf{z}_g)$
        \State Integrate $\widehat{\mathbf{A}}_{t}^{(k)}$ from the current pose to obtain its predicted trajectory
    \EndFor
    \State Compute the discounted latent loss $\mathcal{L}_{\mathrm{WM}}$
    \State Compute $\mathcal{L}_{\mathrm{PLA}}$ by averaging normalized rATE and rRPE over all depths
    \State Update $\theta$ and $\psi$ using $\mathcal{L}_{\mathrm{WAM}}=\lambda_{\mathrm{WM}}\mathcal{L}_{\mathrm{WM}}+\lambda_{\mathrm{PLA}}\mathcal{L}_{\mathrm{PLA}}$
\EndFor
\State Freeze $E$, $R_{\rho}$, $F_{\theta}$, and $P_{\psi}$

\Statex \textbf{Stage III: Gate preference training}
\For{each sample $i$ in $\mathcal{D}_{\mathrm{tr}}$}
    \For{$k=1,\ldots,K_{\max}$}
        \State Generate $\widehat{\mathbf{z}}_{t+1:t+k}$, $\widehat{\mathbf{A}}_{t}^{(k)}$, and the causal Gate input $\mathbf{u}_{i,k}$
        \State $C_{i,k} \gets \mathrm{rATE}_{i,k}/\epsilon_A+\mathrm{rRPE}_{i,k}/\epsilon_R+\lambda_c k/K_{\max}$
    \EndFor
    \State $k_i^{+}\gets\arg\min_k C_{i,k}$
    \State Add $(\mathbf{u}_{i,k_i^{+}},\mathbf{u}_{i,k^{-}})$ to $\mathcal{D}_{\mathrm{pref}}$ for every $k^{-}\neq k_i^{+}$
\EndFor
\State Initialize and freeze the reference Gate $p_{\mathrm{ref}}$ from $p_{\omega}$
\For{each preference minibatch $(\mathbf{u}^{+},\mathbf{u}^{-})$ from $\mathcal{D}_{\mathrm{pref}}$}
    \State Compute $r_{\omega}(\mathbf{u}^{+})$ and $r_{\omega}(\mathbf{u}^{-})$ relative to $p_{\mathrm{ref}}$
    \State Update only $\omega$ by minimizing $\mathcal{L}_{\mathrm{DPO}}$
\EndFor
\State Calibrate the depth-specific thresholds $\{\eta_k\}$ on $\mathcal{D}_{\mathrm{cal}}$
\end{algorithmic}
\end{algorithm}

\clearpage

\begin{algorithm}[h]
\caption{Adaptive DroneWAM Inference}
\label{alg:dronewam_inference}
\footnotesize
\begin{algorithmic}[1]
\Require Observations $\mathbf{o}_{\leq t}$, motion states $\mathbf{s}_{\leq t}$, executed actions $\mathbf{a}_{\leq t}$, goal image $g$, thresholds $\{\eta_k\}$, and depths $K_{\min},K_{\max}$
\Ensure Executed action $\widehat{\mathbf{a}}_{t+1|t}^{(\kappa_t)}$, stopping depth $\tau_t$, and selected-plan depth $\kappa_t$

\State $\mathbf{z}_{\leq t}\gets R_{\rho}(E(\mathbf{o}_{\leq t}))$ and $\mathbf{z}_g\gets R_{\rho}(E(g))$
\State $C_t\gets\{\mathbf{z}_{\leq t},\mathbf{s}_{\leq t},\mathbf{a}_{\leq t}\}$
\State Initialize $\tau_t\gets K_{\max}$ and $\kappa_t\gets1$
\For{$k=1,\ldots,K_{\max}$}
    \State $\widehat{\mathbf{z}}_{t+k}\gets F_{\theta}(C_t,\mathbf{z}_g,\widehat{\mathbf{z}}_{t+1:t+k-1})$
    \State $\widehat{\mathbf{A}}_{t}^{(k)}\gets P_{\psi}(\widehat{\mathbf{z}}_{t+1:t+k},\mathbf{z}_g)$
    \State Construct the causal Gate input $\mathbf{u}_{t,k}$ from $C_t$, $\widehat{\mathbf{z}}_{t+1:t+k}$, and $\widehat{\mathbf{A}}_{t}^{(1:k)}$
    \State $q_{t,k}\gets\sigma\!\left(\beta r_{\omega}(\mathbf{u}_{t,k})\right)$
    \If{$k=1$ or $q_{t,k}>q_{t,\kappa_t}$}
        \State $\kappa_t\gets k$
    \EndIf
    \If{$k\geq K_{\min}$ and $k<K_{\max}$ and $q_{t,k}\geq\eta_k$}
        \State $\tau_t\gets k$; \textbf{break}
    \EndIf
\EndFor
\State Execute only $\widehat{\mathbf{a}}_{t+1|t}^{(\kappa_t)}$, the first action of $\widehat{\mathbf{A}}_{t}^{(\kappa_t)}$
\State Receive the next observation and repeat in a receding-horizon manner
\end{algorithmic}
\end{algorithm}

\end{document}